\documentclass[journal]{IEEEtran}
\usepackage{multirow} 
\usepackage{booktabs}
\usepackage{makecell}
\usepackage{cite}
\usepackage{amsmath} 
\usepackage[ruled]{algorithm2e} 
\usepackage{array}
\usepackage{graphicx}
\usepackage{float}
\usepackage{fixltx2e}
\usepackage{amsthm,amsmath,amssymb}
\usepackage[table,xcdraw]{xcolor}
\graphicspath{{./figures/}}
\usepackage{url}  
\usepackage[ruled]{algorithm2e} 

\ifCLASSOPTIONcompsoc
  \usepackage[caption=false,font=normalsize,labelfont=sf,textfont=sf]{subfig}
\else
  \usepackage[caption=false,font=footnotesize]{subfig}
\fi

\begin{document}
\title{MSSPN: Automatic First Arrival Picking using Multi-Stage Segmentation Picking Network}

\author{Hongtao~Wang, Jiangshe~Zhang, Xiaoli~Wei, Chunxia~Zhang, Zhenbo~Guo, Li~Long, Yicheng~Wang 
\thanks{Corresponding authors: Jiangshe Zhang; Chunxia~Zhang.}
\thanks{H.T. Wang, J.S. Zhang, C.X. Zhang, X.L. Wei, L. Long, Y.C. Wang are with the School of Mathematics and Statistics, Xi’an Jiaotong University, Xi’an, Shaanxi, 710049, P.R.China.}
\thanks{Z.B. Guo are with the Geophysical Technology Research Center of Bureau of Geophysical Prospecting, Zhuozhou, Hebei, 072751, P.R.China.}
\thanks{The research is supported by the National Key Research and Development Program of China under grant 2020AAA0105601, the National Natural Science Foundation of China under grant 61976174 and 61877049.}}

\markboth{Journal of \LaTeX\ Class Files,~Vol.~14, No.~8, August~2015}%
{Shell \MakeLowercase{\textit{et al.}}: Bare Demo of IEEEtran.cls for IEEE Journals}

\maketitle

\begin{abstract}
Picking the first arrival times of prestack gathers is called First Arrival Time (FAT) picking, which is an indispensable step in seismic data processing, and is mainly solved manually in the past. With the current increasing density of seismic data collection, the efficiency of manual picking has been unable to meet the actual needs. Therefore, automatic picking methods have been greatly developed in recent decades, especially those based on deep learning. However, few of the current supervised deep learning-based method can avoid the dependence on labeled samples. Besides, since the gather data is a set of signals which are greatly different from the natural images, it is difficult for the current method to solve the 
FAT picking problem in case of a low Signal to Noise Ratio (SNR). In this paper, for hard rock seismic gather data, we propose a Multi-Stage Segmentation Pickup Network (MSSPN), which solves the generalization problem across worksites and the picking problem in the case of low SNR. In MSSPN, there are four sub-models to simulate the manually picking processing, which is assumed to four stages from coarse to fine. 
Experiments on seven field datasets with different qualities show that our MSSPN outperforms benchmarks by a large margin.
Particularly, our method can achieve more than 90\% accurate picking across worksites in the case of medium and high SNRs, and even fine-tuned model can achieve 88\% accurate picking of the dataset with low SNR.
\end{abstract}

\IEEEpeerreviewmaketitle

\section{Introduction}
\IEEEPARstart{F}{irst} Arrival Time (FAT) picking is an essential step in seismic data processing, and any pickup error can significantly impact subsequent seismic processing\cite{yilmaz2001seismic}. Therefore, accurate FAT picking is an inevitable problem in seismic data processing. With the development of geological exploration acquisition technology, the data acquisition density is extremely high. Manual picking work is inefficient and subjective so that first arrival picking only manually is an impossible task in the face of high-density data. Specifically, when the low-quality and high-density pre-stack data is manually picked for the first arrivals, the processing time accounts for 20\%-30\% of the total data processing flow\cite{sabbione2010automatic}. In recent decades, an amount of automatic FAT picking methods have been proposed, and gradually replaced manual picking processing. 

In earlier studies, seismic wave FAT picking problem boils down to a task to identify mutation points based on the statistics of a single trace (called single-level) or adjacent traces (called multilevel). Jubran et al.\cite{akram2016review} summarize some classical traditional FAT picking methods and classify the methods based on the scale of input data as: window-based single-level methods, nonwindow-based single-level methods, multilevel- or array-based methods, and hybrid methods in which the above methods are combined and applied to the picking processing. 
Concretely, there are lots of energy ratios or high-order statistics for recognizing the first abnormal point of each time series, such as, Short-Time Average and Long-Time Average ratio (STA/LTA)\cite{allen1978automatic}\cite{baer1987automatic}, Modified Energy Ratio (MER)\cite{han2009time}, Akaike Information Criterion (AIC)\cite{takanami1991estimation}\cite{sleeman1999robust}, skewness and kurtosis\cite{yung1997example}\cite{saragiotis2004automatic}, etc. Moreover, the data can also be transformed firstly to a new domain to split the signals and noises better, such as, ray-centered coordinates\cite{akram2016review}, projection-onto convex-sets\cite{abma20063d}, synchrosqueezed continuous wavelet transform\cite{mousavi2016automatic}, discrete Hilbert transform\cite{souza2017automatic}), etc. Obviously, the single-level methods lose the using of spatial information, which causes the instability of the picking results. In addition, although the multi-level methods or hybrid methods alleviate the instability of the picking results in some degree, it still seems powerless in case of median or low SNRs. 

Recently, increasing number of machine learning and deep learning methods are applied to the FAT picking task, such as, classification-based methods, semantic segmentation-based methods, regression-based methods, reinforcement learning-based methods, active learning-based methods, etc. In 1990s, neural networks are used as classifiers to identify the locations of FAT roughly on each single trace\cite{mccormack1993first}\cite{dai1997application}. Recently, Xu et al.\cite{duan2018integrating} construct two Convolutional Neural Networks (CNNs) to identify the poor picks across multiple traces and correct them, respectively. Yaniv et al.\cite{hollander2018using} and San et al.\cite{yuan2018seismic} first resort a CNN to recognize the coarse FAT locations, and then utilize the other conventional methods to fine the accurate FAT. Moreover, in order to perceive easily, there is a very common practice of generating multiple features of the signal data as input to a regression learner or a classification learner to achieve a gain on the signals of FAT\cite{maity2014novel}\cite{chen2019first}\cite{duan2019multi}\cite{guo2020aenet}. 

Different from the above classification-based methods, Fei et al.\cite{luo2018automatic} develop a method of FAT picking based on instantaneous traveltime constrained by conventional image segmentation. In order to alleviate the lack of labeled samples, Kuo et al.\cite{tsai2018first}\cite{tsai2019automatic} propose a semi-supervised segmentation network to improve model perception. However, the FAT picking task has extremely high requirement on picking accuracy, and it causes that the segmentation mask is very sparse in this case. Fortunately, current medical image segmentation can solve the unstable segmentation problem caused by the sparse mask. As a classic method for medical segmentation networks, U-Net is first applied to the FAT picking task in 2019\cite{hu2019first}\cite{ma2019automatic}, and then nested U-Net\cite{zhou2018unet++}, a combination of U-Net and Dense-Net\cite{huang2017densely}, is used to enhance auto-picking generalization\cite{zhang2020first}. Subsequently, in order to ensure the stability of the output results of machine learning-based methods, some post-processing methods based on velocity constraints are proposed\cite{cova2020automated}\cite{liu2020technology}. Recently, Pierre et al.\cite{P2021Neurips} have open-sourced a set of public datasets with manual annotations and provide a U-Net-based pickup accuracy benchmark, which we will also compare in our experiment section. 
Apart from the above methods, there are some other methods to assist the machine learning-based first arrival picking method. Ma et al.\cite{ma2018automatic} utilize the reinforcement method to pick the FAT by trace-wise. Xie et al.\cite{xie2019first} attempt to use the transfer learning method to improve the generalization performance of the trained model. Richardson\cite{richardson2021active} propose a sample selection method based on active learning to improve dataset quality. 

Based on the above description of automatic picking methods, we briefly summarize the challenges faced by various methods. 
First, it is hard for traditional methods or machine learning-based picking methods based on regression or classification to pick FAT stably in case of medium or low SNRs, caused by only using the amplitude information of a single trace or a local area to identify the first arrival time. 
Second, since the decision boundary is usually sparse, current segmentation-based picking methods cannot recognize patterns near FAT on the dataset with low SNR, so that it greatly restricts the generalization across worksites. 
Third, to the best of our knowledge, current deep learning-based picking method ignores the other gather information, such as the offsets between sources and receivers. 
In this paper, we propose a new FAT picking method named Multi-Stage Segmentation Picking Network (MSSPN) to solve the above three problems. We summarize our contributions as follows.

\begin{enumerate}[\IEEEsetlabelwidth{4}]
	\item We propose a new artificial intelligence-based automatic picking framework, which imitates the manual picking process with multiple stages of coarse-to-fine picking. 
	\item In our method, we embed linear moveout (LMO) correction into the picking process, means that, more geophysics knowledge is introduced to improve the robustness of semantic segmentation-base method.
	\item We develop a constrained parameter inference method to estimate the parameters of LMO, instead of traditional manual setting.
	\item We also propose a new robust post-processing method to remove the uncertain FAT picking points.
\end{enumerate}

\section{Methodology}
\begin{figure*}[htbp]
	\centering
	\includegraphics[width=6in]{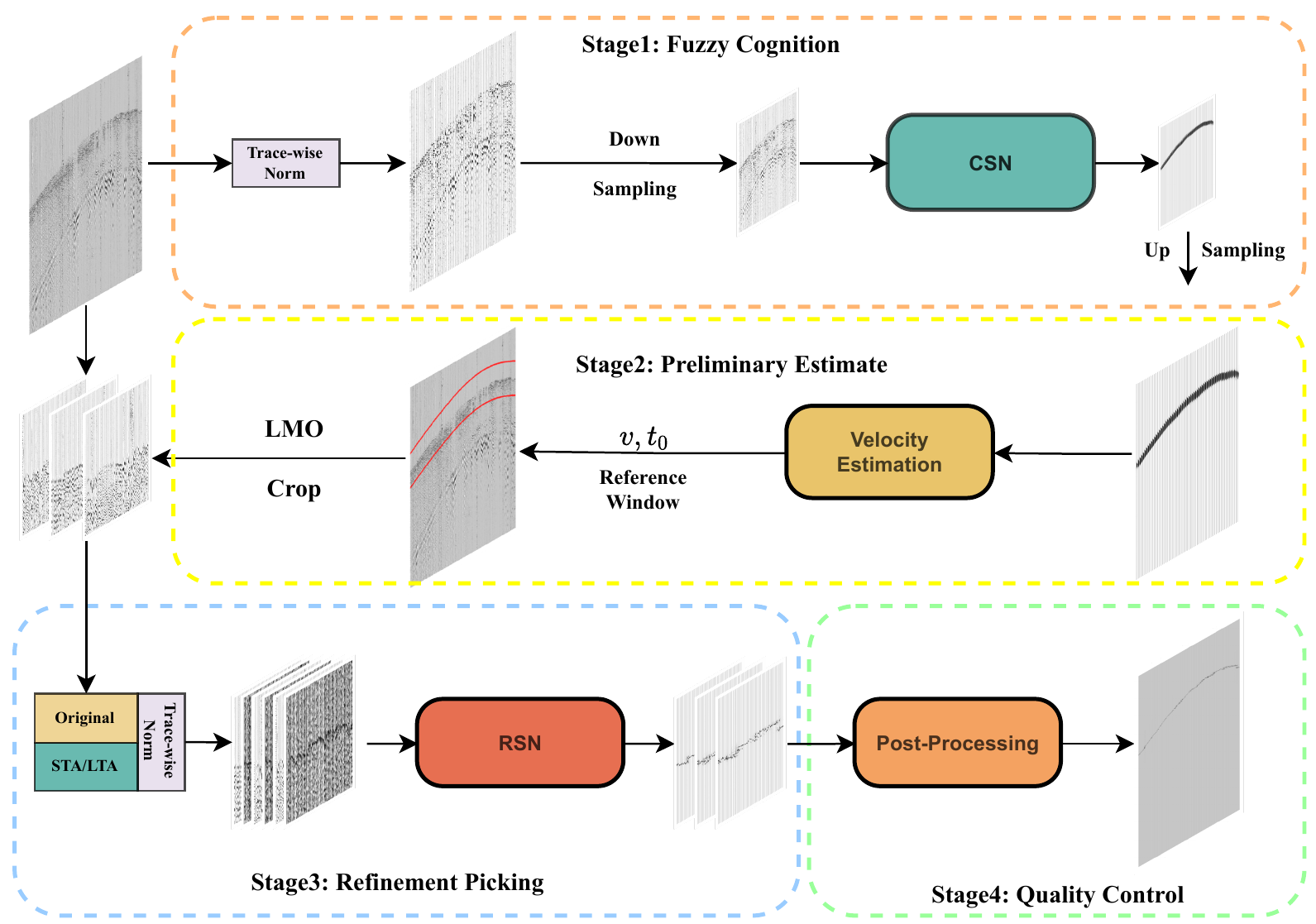}
	\caption{The flowchart of MSSPN}
	\label{fig: MSSPN}
\end{figure*}

To further improve the accuracy and stability of picking in case of medium and low SNRs, we propose a new picking framework as shown in Fig. \ref{fig: MSSPN}, in which four stages are developed to imitate the manual picking processing. 
First, Coarse Segmentation Network (CSN) in Stage1 is used to perceive the approximate location of FAT. Second, velocity prior is utilized to estimate robust intervals of FAT based on the prediction map of CSN. Third, Refine Segmentation Network (RSN) in Stage3 is resorted to finely pick the first arrival times on a smaller interval, which is obtained by Stage2. Finally, a post-processing model ensures stable picking results. In this section, we will introduce the main ingredients of MSSPN in order.

\subsection{Data Pre-processing}
Prestack gather data is a time series signal arranged in order based on specific properties, and its amplitude, period, and frequency are all very important features of each single trace. In order to better improve the pick-up rate and reduce the pick-up error rate, we preprocess the signal data and label data separately.

First, we introduce the preprocessing of the signal data. To enhance the local perception of first arrival time, we generate the STA/LTA ratio\cite{allen1978automatic} feature map, in which STA/LTA ratio can greatly highlight the boundaries of all energy abrupt changes, using the following equation on each single trace:
\begin{equation}
	\text{STA/LTA}_i = {{{n_l}\sum\limits_{j = i - {n_s}}^i {\left| {\text{amp}}_j \right|} } \over {{n_s}\sum\limits_{j = i - {n_l}}^i {\left| {\text{amp}}_j \right|} }},
	\label{STALTA}
\end{equation}
where $\text{STA/LTA}_i$ and ${\text{amp}}_i$ is STA/LTA ratio value and amplitude value of i-th element in a single trace, $n_s$ and $n_l$ are the length of short window and long window, respectively. Compared with original gather, STA/LTA feature map enhances the perception of the possible first break points. Then, we normalize the signal data trace-by-trace named trace-wise normalization to ensure the consistency of the data distribution of the input:
\begin{equation}
  {\text{amp}}_i' = {{\text{amp}}_i \over {\mathop {\max }\limits_{1 \le j \le N} \left\{ {{\text{amp}}_j \cdot Sign\left( {\text{amp}}_j \right)} \right\}}},
  \label{TWNorm}
\end{equation}
where ${\text{amp}}_i'$ and ${\text{amp}}_i$ is the normalization result and amplitude value of i-th element in a single trace, $Sign(\cdot)$ is the signnum function, and the $N$ is the number of the elements in a single trace.

Second, we also calibrate the human-labeled FAT of prestack gather data. Due to the different ways of collecting data, the optimal pick-up positions of FAT are different. Concretely, in the vibroseis acquisition method, FAT should be marked at peaks, while the explosion source is marked at jumping-off points. Most of the manual picking are obtained by experts correcting the picking results of STA/LTA method, so the position of the mark is not exactly on peaks or troughs. For deep learning models, accurate patterns are very necessary, for example, marking the peaks (troughs), that means, we move the labels of non-peak (non-trough) times to the nearest peak (trough) times. The inconsistency of the labeling pattern (peaks or troughs) often leads to overfitting of the deep learning model training, and we will also verify it in the experimental part.

\subsection{Multi-Stage Segmentation Picking Network (MSSPN)}
To best of our knowledge, in the current automatic FAT picking models, the supervised deep learning-based methods can identify the FAT position well, but the generalization is weak, that is, it is difficult to recognize across work sites (data domains). To pick the FAT in case of medium or low SNRs steadily across the work sites, we propose the MSSPN.

We assume that there are four main stages of recognition in the manual picking processing:
\begin{enumerate}[\IEEEsetlabelwidth{3}]
	\item Stage 1, fuzzy perception: analysts observe roughly and obtain the approximate positions, where there may be FATs.
	\item Stage 2, preliminary estimate: analysts preliminarily estimate the pickup range based on the fuzzy perception results.
	\item Stage 3, refinement picking: in a small area obtained by Stage2, analysts carefully identify the specific location of FAT. 
	\item Stage 4, quality control: analysts remove less confident pickups. 
\end{enumerate}

So, our proposed MSSPN has four sub-models as shown in Fig. \ref{fig: MSSPN}, corresponding to the four stages in the above assumptions. First, we give a coarse segmentation probability map based on the down-sampled image of original gather. Secondly, based on the coarse segmentation results and offsets between sources and receivers of each trace, we estimate the velocity ($v$) and zero-offset traveltime ($t_0$), which are the two main parameters of LMO. Then, we determine the approximate pick-up range using the LMO correction. Next, the mixed feature maps based on the full-resolution gathers of small regions give the more accurate segmentation results. Finally, we make corrections to the refine-segmentation results based on a post-processing method. 

In principle, two segmentation networks in stages 1 and 3 of MSSPN can be replaced by any semantic segmentation network. In this paper, we choose U-Net as the segmentation network in MSSPN. U-Net\cite{ronneberger2015u} was proposed in 2015 to solve the problem of medical image segmentation, which has been widely used in seismic data processing, e.g. velocity auto-picking\cite{wang2021automatic}, seismic data reconstruction\cite{park2021method}, seismic data interpolation\cite{fang2021seismic}, first break\cite{hu2019first}, etc. It is composed by a few Down-sample Blocks (DBs) and some Up-sample Blocks (UBs). To keep the original semantic information and avoid vanishing gradient, skip connection is implemented between the same depth of encoder and decoder. DB consists of two convolution-batch normalization-ReLU (CBR) module and a max pooling layer. CBR module consists of a convolutional layer with kernel size = $3 \times 3$, stride size = $1 \times 1$, padding size = $1 \times 1$, a batch-normalization layer, and a ReLU  activation layer.
In our work, the up-sampling part of UB is a 2D transposed convolution operator with kernel size = $4 \times 4$, stride size = $2 \times 2$, padding size = $1 \times 1$, and then input the ReLU activation layer. After concatenating the original feature maps and the up-sampled feature map, finally there are two CBR modules. At last, there are a convolution layer with kernel size = $3 \times 3$, stride = $2 \times 2$, padding size = $1 \times 1$ and a Sigmoid activation layer. Thus, the output of the U-Net is a probability segmentation prediction map, which means that the value of each element is between 0 and 1. Next, we will introduce our four sub-models in detail.

\subsubsection{Coarse-Segmentation Network (CSN)}
\begin{figure}[ht!]
	\centering
	\subfloat[]{\includegraphics[height=1.5in]{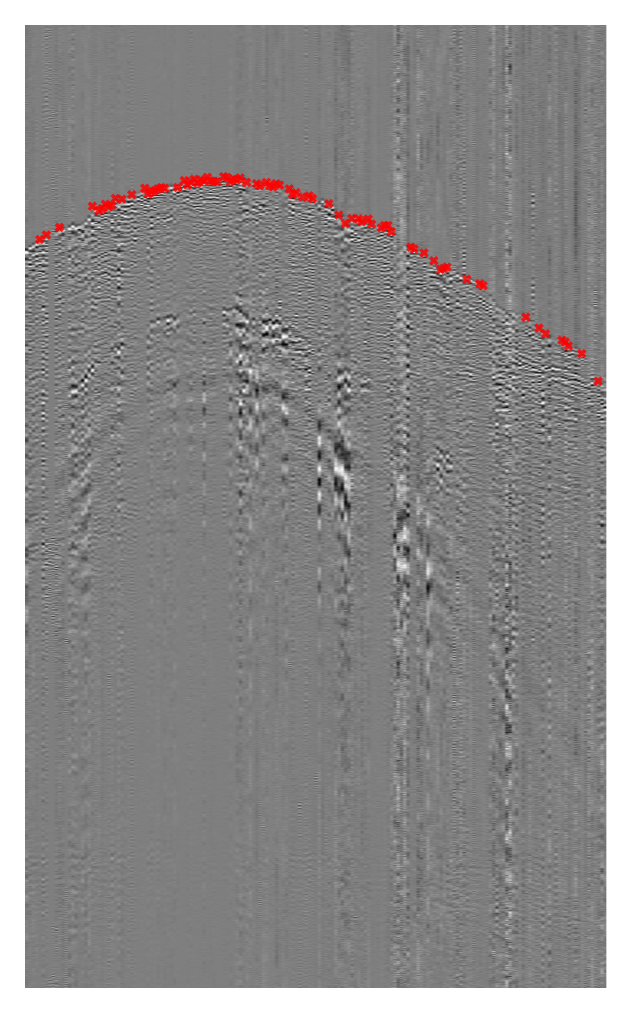}\label{fig: Ori}}
	\hfil
	\subfloat[]{\includegraphics[height=1.5in]{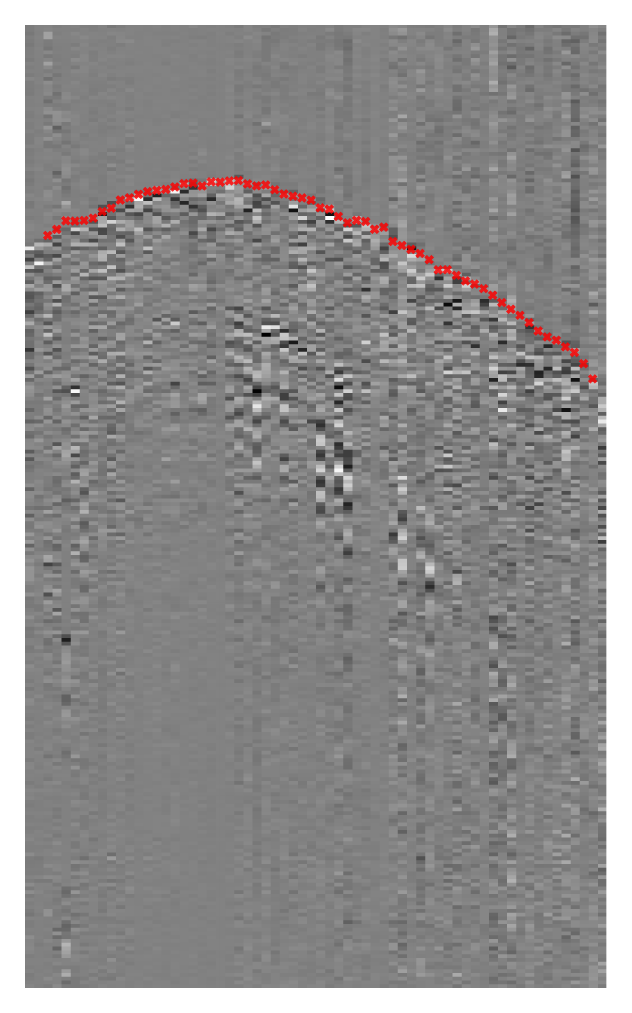}\label{fig: DS}}
	\hfil
	\subfloat[]{\includegraphics[height=1.5in]{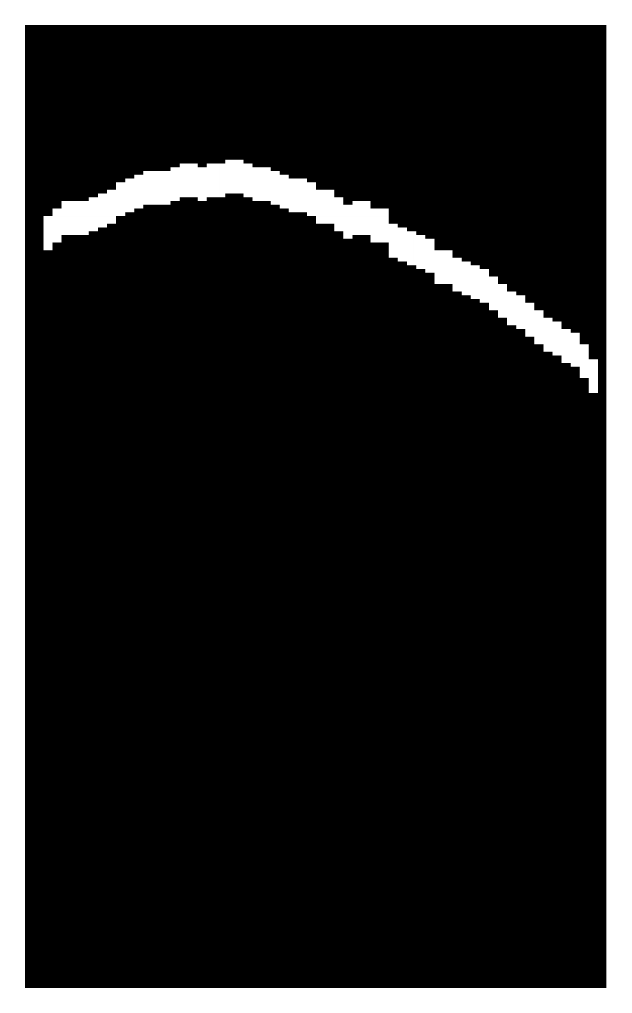}\label{fig: GT}}
	\caption{(a) Original gather grey image with manual first arrival label. (b) Downsampled gather grey image with interpolated label. (c) The ground-truth image of (b) in CSN.}
	\label{fig: DownSampling}
\end{figure}

Stacking gather data to improve imaging quality is an important method in seismic data processing, which essentially observes data on a large scale (macro) to achieve the effect of denoising. In deep learning, there are two approaches to increase the receptive field. One is increasing the size of the convolution kernel, and the other is down-sampling the image. In our paper, we use the downsampled subgraphs as the input of CSN, which enables to increase the relative receptive field range of the model with reasonable computational complexity. 
Another reason to use downsampled sub-images as input is that for full-resolution gather images, the labels of the first arrival times are very sparse. For example, on a signal of length 1000, only a pixel is labeled as 1. 
If a simple model is used to recognize, it is easy to overfit. In turn, if a model with high complexity is used for training, a large amount of labeled data is required to train the model. Neither of these are sensible, thus, we settle for the next best thing, hoping to use the downsampled subgraphs to identify a rough FAT trajectory.
Although the downsampled image does lose local high-frequency information, it can be seen from Fig. \ref{fig: DownSampling}
that the approximate first arrival time trend can be observed in the downsampled image. Therefore, it is reasonable to downsample the image as the input of the CSN. 

In CSN, the shape of the input image is $1\times H \times W$, and the network structure adopts a U-Net with depth $=4$, in which there are four pairs of DB and UB, and each pair of DB-UB has a skip connection, and the shape of the final output (the segmentation probability map) is $H \times W$. Before input to network, the gather image is normalized using trace-wise normalization, and is downsampled subsequently to shape = $H \times W$ as the input of the CSN.
Moreover, since the precision required for rough picking is not high, we interpolate the manual labels using linear interpolation method as shown in Fig. \ref{fig: DS}, and make the corresponding ground-truth as shown in Fig. \ref{fig: GT}, where the nearby 5 pixels of true FAT on each trace are also marked as 1. 

\subsubsection{Regional Restriction based on Velocity Estimation (RRVE)}
CSN has provided a rough segmentation probability map, but there is still a lot of noise. Therefore, we propose a regional restriction method based on velocity estimation to remove the noise. Although there have been some works based on region constraints to narrow the input range, such as \cite{xu2020high}, these intervals or local velocities are all artificially given or estimated by labeled seeds, which is contrary to the idea of full automation. Generally, the offsets between the source point and the receiver points during seismic data acquisition are not large, so the local surface velocity is relatively constant. A rough velocity can be estimated to determine the approximate range of FAT based on LMO. To estimate a reasonable local velocity and zero-offset traveltime, we develop a RANdom SAmple Consensus algorithm \cite{fischler1981random} with Vlocity Constraints (RANSACwVC).

Each trace on the gather has corresponding coordinates of the receiver point and the source point, so the distance between the two can be calculated. We first use a splitting threshold to separate the points with high probability in the rough segmentation probability map, and transform their corresponding horizontal and vertical coordinates into the time-offset domain, denoted as point set $\left\{ {\left( {{d_i},{t_i}} \right)} \right\}_{i = 1}^N$. Under the assumption of constant local velocity, point set $\left\{ {\left( {{d_i},{t_i}} \right)} \right\}_{i = 1}^N$ can be fitted by the equation of LMO:
\begin{equation}
  t = d / v + t_0
  \label{Linear}
\end{equation}
where $v$ is the constant local velocity. When the current region has prior knowledge of velocity, we can constrain the estimated constant local velocity, and then construct the following optimization model to solve the velocity $v$ and zero-offset time $t_0$:
\begin{equation}
  \begin{array}{c}
    {\mathop {\text{minimize} }\limits_{{t_0},v} \left\{ {\sum\nolimits_i {{{\left[ t_i - ( d_i / v + t_0 ) \right]}^2}} } \right\}} \\
    {\text{subject to}\ {V_{\min }} < v < {V_{\max }},\ t_0 > 0},
  \end{array}
  \label{OptRAN}
\end{equation}
where, $V_{\min }$ and $V_{\max }$ are the upper and lower bounds of the velocity, respectively. The above optimization problem can be solve by Fortran subroutines for large-scale bound-constrained optimization (L-BFGS-B)\cite{zhu1997algorithm}.
\begin{figure}[ht!]
	\centering
	\subfloat[]{\includegraphics[width=0.5\linewidth]{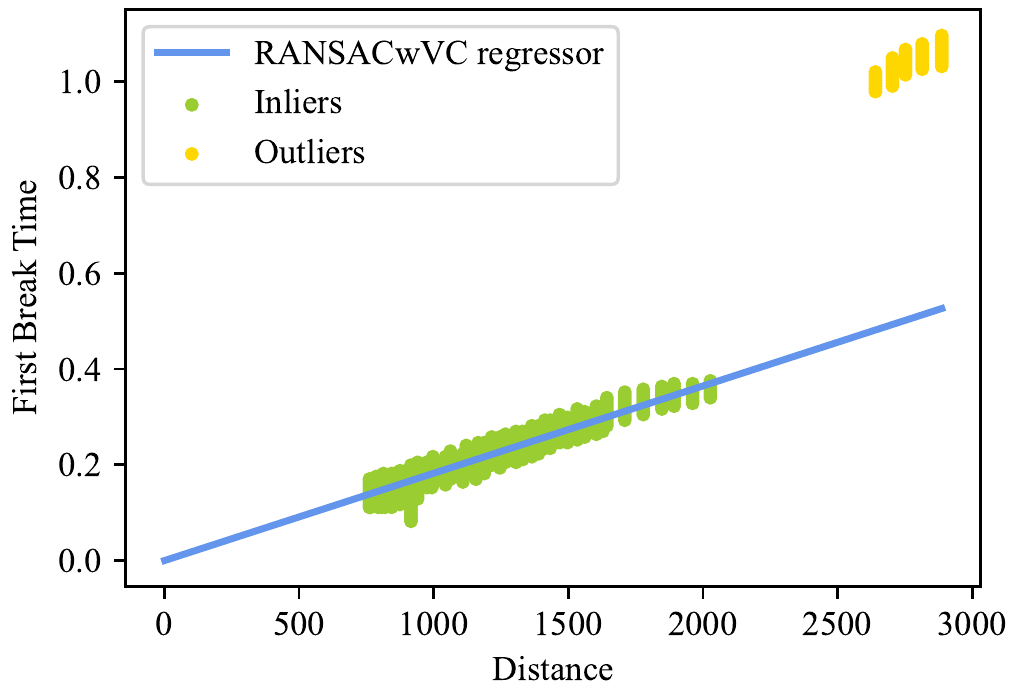}\label{fig: RANSCA-1}}
	\hfil
	\subfloat[]{\includegraphics[width=0.5\linewidth]{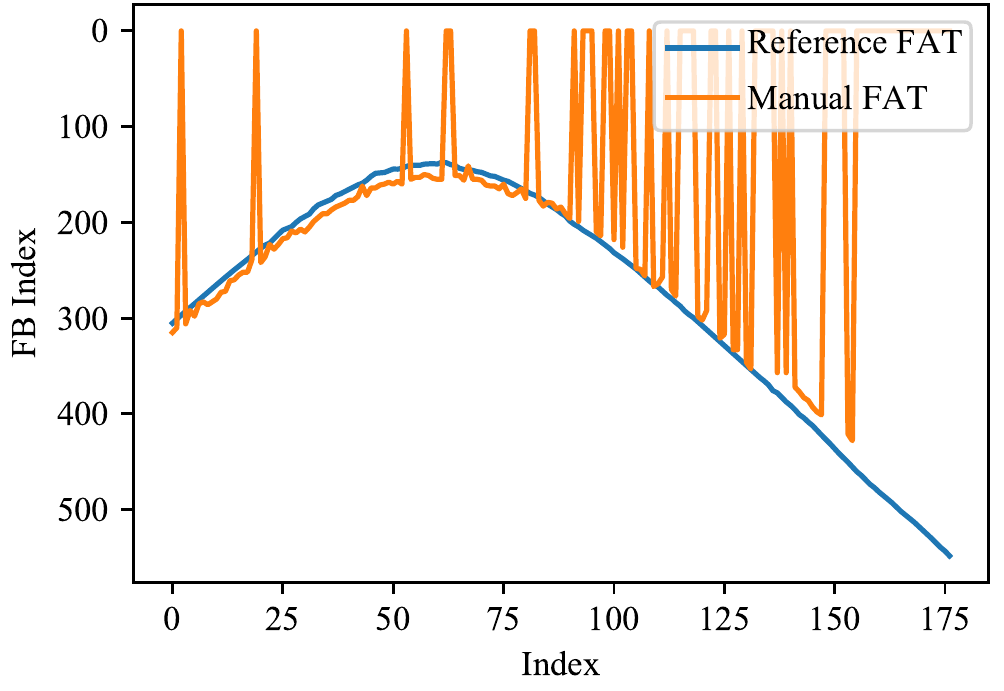}\label{fig: RANSCA-2}}
	\\
	\subfloat[]{\includegraphics[height=1.55cm]{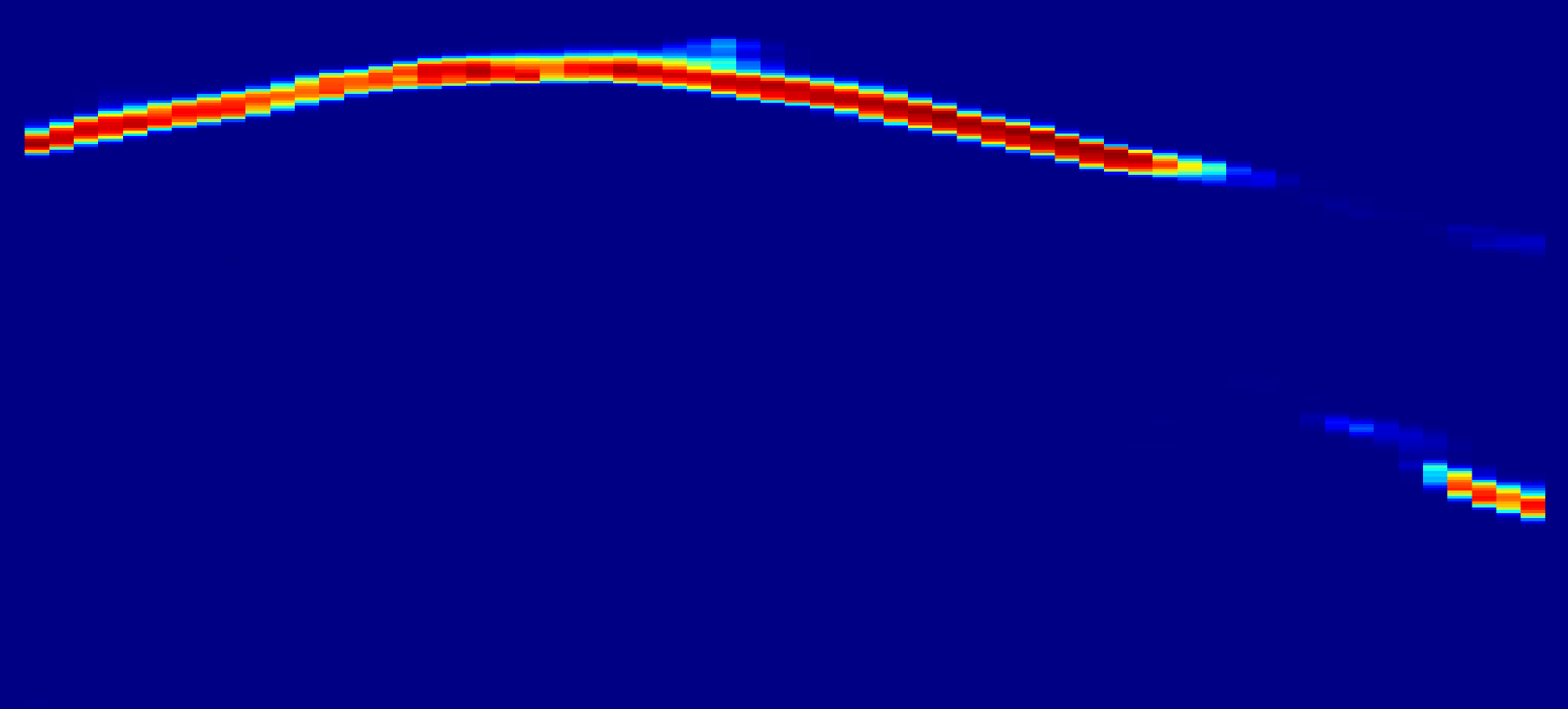}\label{fig: LMO}}
	\hfil
	\subfloat[]{\includegraphics[height=1.55cm]{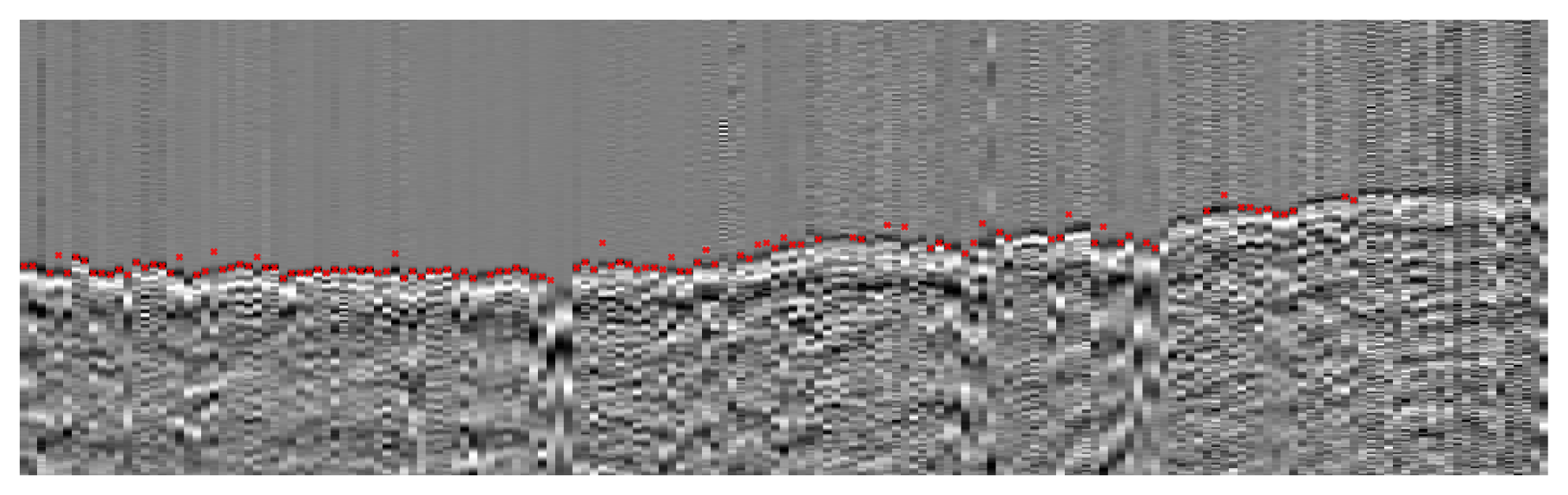}\label{fig: LMO}}
	\caption{(a) RANSACwVC regression processing. (b) Manually picked first arrival times and the reference first arrival times estimated by RANSACwVC. (c) The prediction map of CSN. (d) LMO gather with manually picked first arrival times (red points).}
	\label{fig: RANSAC}
\end{figure}

RANSAC is a robust iterative algorithm for correctly estimating the parameters of a mathematical model from a set of data containing outliers, and the brief algorithm description as shown in Algorithm. \ref{ALG:RANSAC}. RANSAC has stronger robustness, that means, for instance, it can effectively eliminate the interference of outliers compared to the direct least squares method for model parameter estimation in linear regression model. Although there is already a method of picking up the first arrival time using RANSAC, the difference is that Yoo et al.\cite{yoo2020line} is a piece-by-piece linear approximation in the original gather, and we perform denoising regression on the linear function on the time-offset domain as shown in Fig. \ref{fig: RANSCA-1}, which is more in line with the assumptions of the physical model. After solving the Eq.\ref{OptRAN} by RANSAC, we can calculate the reference first arrival time for each trace based on estimated parameters ${\hat v}_b, {\hat t}_{{b_0}}$, as show in Fig. \ref{fig: RANSCA-2}, using the following equation:
\begin{equation}
  {{\tilde t}_i} = {d_i}/{{\hat v}_b} + {{\hat t}_{{b_0}}}.
  \label{PredT}
\end{equation}
However, since the surface velocity is not a constant value, we cannot directly use the reference first arrival time as the final results, but only as the center position of a reference window. Therefore, we set a confidence interval near the reference first arrival time to ensure that the interval can cover most of the first arrival times. In addition, we can observe from Fig. \ref{fig: LMO} that the first arrival time is almost in the state of a horizontal line, which greatly reduces the training difficulty of the subsequent Refine-Segmentation network.

\begin{algorithm}[htbp]
	\caption{RANSACwVC Algorithm}
	\KwIn{samples $S_0=\{(d_i, t_i)\}^N_{i=1}$, model $M$, minimum sample number of model need $n$, minimum sample number of manual need $d$, iteration time $k$, judgement threshold $t$}
	\KwOut{best model parameter ${\hat v}_b,{\hat t}_{b_0}$\;}
	Initialize iteration time  $iter = 0$\;
	Initialize best consensus set $S_c = \emptyset$\;
	Initialize best error $ E_B =  + \infty$ \;
	\While{iter $< k$}{
	    $S_{rs} = $ randomly select n points from the set $S_0$\;
	    $\hat v,{{\hat t}_0} = $ solve optimization model $M$ (Eq.\ref{OptRAN}) based on set $S_{rs}$\;
	    \For{$(d_j, t_j)$ in $S_0\backslash S_{rs}$}{
	    $l_j = {{{\left[ t_j - ( d_j / {\hat v} + \hat{t}_0 ) \right]}^2}} $\;
	    \If{$l_j < t$}{Add $(d_j, t_j)$ into consensus set $S_c$\;}
	    }
	    \tcp{\small{$\#(\cdot)$: count element number}}
	    \If{$\#(S_c) > d$ }
	    {
	        ${\hat v}_n,{\hat t}_{n_0} = $ solve optimization model $M$ (Eq.\ref{OptRAN}) based on set $S_{c}$\;
	        $L = \sum\nolimits_{{S_c}} {\left[ {{t_i} - \left( {{d_i}/{{\hat v}_n} + {{\hat t}_{{n_0}}}} \right)} \right]^2} $\;
	        \If{$L < E_B$}{
	        ${\hat v}_b,{\hat t}_{b_0} = {\hat v}_n,{\hat t}_{n_0}$\;
	        $E_B = L$
	        }
	    }
	    $iter ++$
	}
	\Return{${\hat v}_b,{\hat t}_{b_0}$}
	\label{ALG:RANSAC}
\end{algorithm}

\subsubsection{Refine-Segmentation Network (RSN)}
To further improve the picking accuracy, we design RSN based on U-Net. After RRVE, we get a smaller region, which covers all first arrival times. 
Before introducing the network structure, we first describe the data preparation. First, we crop horizontally the LMO image (e.g. Fig. \ref{fig: LMO}) into a few mini-patches with the same shape ${H'}\times {W'}$. Second, we generate the feature map of STA/LTA corresponding to these mini-patches to enhance local features near the first arrival time. Finally, we utilize trace-wise normalization on the original mini-patch and its corresponding STA/LTA feature map, forming an array of shape $2 \times {H'}\times {W'}$ as the input of RSN.
In RSN, there are two differences from the network structure of CSN. First, we choose U-Net with a depth of 5 in this paper to build a more powerful network. Second, we only label the pixel corresponding to the manual first arrival time in the ground-truth map as 1.

\subsubsection{Post-processing}
Although our refined picking process is carried out in a small interval, which has avoided serious misalignment picking, it is well known that the output results of deep learning models are often not robust. Therefore, we still need to impose other prior information to constrain the output results. We adopt a similar approach to RRVE to remove partial pickups, which do not conform to physical assumptions.
Our post-processing is divided into four steps. 
First, the prediction maps of the mini-patches in the RSN are spliced into a prediction map of the whole gather, and we will average the overlapping parts.
Second, we filter out the maximum value of each column from the RSN prediction map as the selected candidate point set $P_0$. Third, a reference pick-up is estimated using RRVE, which is similar to the blue curve in Fig. \ref{fig: RANSCA-2}. 
Finally, we remove the picking points in $P_0$ whose deviation from the reference picking is greater than $T_d$, and the remaining points are considered as the final first-arrival points. We recommend setting the deviation threshold $T_d$ in the range of 5-20 pixel, and in our paper it is 5 pixel.

\subsection{Quality Control}
In this subsection, we will introduce a few measure methods to evaluate the quality of automatic picking. To facilitate the understanding, we denote the manual picking and the automatic picking as $\left\{ {{P_{{M_i}}}} \right\}_{i = 1}^N$ and $\left\{ {{P_{{A_i}}}} \right\}_{i = 1}^N$, respectively. Additionally, we mark first arrival times of the no-pickup traces as $-1$ for both manual and automatic picking. Referring to the metrics in \cite{P2021Neurips}, we adapt two classes of measure methods: the picking accuracy and the deviation. The first class is the picking accuracy below k pixel error, denoted as $\text{ACC}@k$, which is the accuracy of picking error within k pixels in the traces picked both manually and automatically:
\begin{small}
\begin{equation}
  \text{ACC}@k = {{\sum\nolimits_i {I\left\{ {\left| {{P_{{A_i}}} - {P_{{M_i}}}} \right| < k}, {{P_{{A_i}}},{P_{{M_i}}} > 0} \right\} }} \over {\sum\nolimits_i {I\left\{ {{P_{{A_i}}},{P_{{M_i}}} > 0} \right\}} }},
  \label{ACCK}
\end{equation}
\end{small}
In this paper, we setting $k=1, 3, 9$ for $\text{ACC}@k$ to test the performance under different accuracy requirements. The other class is the deviation, including the Mean Absolute Error (MAE) and the Root Mean Square Error (RMSE), defined by the following equation:
\begin{small}
\begin{equation}
  \text{MAE} = \sum\nolimits_i {\left| {{P_{{A_i}}} - {P_{{B_i}}}} \right|/N},
  \label{MAE}
\end{equation}
\begin{equation}
  \text{RMSE} = \left[\sum\nolimits_i {{{\left| {{P_{{A_i}}} - {P_{{B_i}}}} \right|}^2}/N}\right]^{1/2},
  \label{RMSE}
\end{equation}
\end{small}
where $N$ is the number of all traces. 
\section{Experiments}
\subsection{Datasets}
To evaluate the performance of our proposed method, we select two groups of field datasets. The first group is an open source field dataset\cite{P2021Neurips}, in which there are four datasets named Lalor, Brunswick, Halfmile, and Sudbury, which are all unique mining sites in three provinces of Canada. 
The second group contains three field datasets named DN-1, DN-2, and LS, which are all collected in the mountains of China and provided by the Bureau of Geophysical Prospecting (BGP).
According to the analysis of BGP experts, datasets of Lalor, Brunswick, Halfmile, Sudbury, DN-1, and DN-2 are equipped with medium or high SNRs, which also contain some low SNR data, and LS is a dataset with low SNR. In this paper, a gather sample is a gather of trace, which corresponds to a source and a series of receivers. 
As shown in Tab. \ref{tab:DataInfo}, we focus on Shot Num, Pick $K$+, Pick Rate and Mode, where Shot Num means the number of gather samples, Pick $K$+ means the number of shots whose manually picked trace number are greater than $K$, Pick Rate means the manual annotation rate which is the ratio between the number of manual labeled traces and the number of all traces, and Mode means the mode of FAT (peak or trough).
In these datasets, we focus on the information of the number of gather samples, the number of labeled samples and the manual annotation rate. Moreover, the annotations of each trace of these datasets are first roughly estimated by STA/LTA method, and are subsequently corrected by experts. 
\begin{table}[h]
\centering
\caption{The Basic Information of Datasets}
\label{tab:DataInfo}
\resizebox{\linewidth}{!}{
\begin{tabular}{lccccc}
\toprule\toprule
\textbf{Set Name} & \textbf{Shot Num} & \textbf{Pick 0+} & \textbf{Pick 64+} & \textbf{Pick Rate(\%)} & \textbf{Mode} \\
\hline
Lalor     & 14455 & 12119 & 9213  & 46.13 & trough \\
Brunswick & 18475 & 18457 & 17884 & 83.54 & trough\\
Halfmile  & 5520  & 5502  & 5478  & 90.35 & trough\\
Sudbury   & 11420 & 5106  & 711   & 12.90 & trough\\
DN-1      & 59739 & 58856 & 56134 & 43.34 & peak\\
DN-2      & 45258 & 43988 & 39292 & 39.10 & peak\\
LS        & 2334  & 1778  & 1324  & 27.37 & peak\\
\bottomrule\bottomrule
\end{tabular}}
\end{table}

\subsection{Implementation Details}
In this subsection, we explain the implementation details of MSSPN, which include the input and output of three stages, and introduces the hyper-parameter selection processing. In our work, we use a computer equipped with a RTX 2080 Ti graphic card and a RTX 3090 graphic card for experiment. Moreover, in all the following model training process, we set 20000 as the maximum number of the batch iteration, and validate the model per 15 iterations. We adopt the strategy of early stop to train the model, which breaks the training iteration when the loss of validation processing does not drop eight times consecutively.

In the training of CSN, the input gathers need to be down-sampled by using Bilinear interpolation, then are input into a U-Net network. Finally, U-Net output a probability map. We adopt the Binary Cross Entropy (BCE) loss function in CSN as:
\begin{equation}
  {L_{\text{BCE}}}\left( {p,\hat p} \right) =  - p \cdot \log \hat p - (1 - p) \cdot \log (1 - \hat p).
  \label{BCE}
\end{equation}
In the regional restriction method, we set the splitting threshold as 0.1 to separate the credible points, and the judgement threshold of RANSACwVC is the median of the absolute error. For training RSN, we conduct a LMO correction on the gather based on restricted region, in which the crop window is generated by the estimated $v$ and $t_0$ by Algorithm \ref{ALG:RANSAC}. Because of the enhanced lateral continuity of FAT, we choose a mixed loss function as:
\begin{equation}
   {L_{\text{Mix}}} = {L_{\text{BCE}}} + \lambda  \cdot {L_{\text{grad}}},
  \label{S3Loss}
\end{equation}
where $\lambda$ is a weighting factor and we adopt $\lambda = 0.2$. $L_{grad}$ is the gradient loss based on Sobel operator\cite{irwin1968isotropic}, which is used to find the approximate absolute gradient magnitude as:
\begin{equation}
  \begin{array}{cc}
  {L_{\text{grad}}}\left( {p,\hat p} \right) = \left\| {\text{sobel}\left( p \right) - \text{sobel}\left( {\hat p} \right)} \right\|,    \\ 
  with\ \text{sobel}\left( p \right) = \sqrt {G_x^2\left( p \right) + G_y^2\left( p \right)},\\
  \end{array}
  \label{GradLoss}
\end{equation}
where $G_x$ and $G_y$ are the horizontal and vertical gradient convolution operators, respectively. 

To find suitable hyperparameters, we conduct test experiments with different hyperparameters in Tab. \ref{tab:para-sel} using the grid searching method, and the hyperparameters in bold are the best combinations. Concretely, we train the models using these hyper-parameter combination with dataset setting Fold-4 in Tab. \ref{tab:fold-set}. Each experiment is repeated 10 times using different initialization network weights to avoid the randomness of the experiment, and we select the final appropriate combination based on the average MAE of the test set. 
\begin{table}[h]
\centering
\caption{Hyper-parameter Selection}
\label{tab:para-sel}
\resizebox{0.9\linewidth}{!}{
\begin{tabular}{ccc}
\toprule\toprule
Sub-model & Hyper-paramters         & Setting            \\ \hline
\multirow{5}{*}{CSN} & Down-sample Shape & \pmb{$256\times 64$}, $256\times 128$\\
          & Optimizer               & \pmb{Adam}, SGD          \\
          & U-Net   Depth           & 3, \pmb{4}, 5            \\
          & Initial   Learning Rate & 0.001, \pmb{0.005}, 0.01 \\
          & Training   Batchsize    & 32, \pmb{64}, 128        \\ \hline
\multirow{5}{*}{RSN} & Crop Shape        & \pmb{$256\times 64$}, $256\times 128$                                \\
          & Optimizer               & \pmb{Adam}, SGD          \\
          & U-Net   Depth           & 3, 4, \pmb{5}            \\
          & Initial   Learning Rate & \pmb{0.001}, 0.005, 0.01 \\
          & Training   Batchsize    & 32, \pmb{64}, 128        \\ \bottomrule\bottomrule
\end{tabular}}
\end{table}
\begin{table}[h]
\centering
\caption{Dataset Splitting Setting}
\label{tab:fold-set}
\begin{tabular}{cccc}
\toprule\toprule
\textbf{Fold No.} & \textbf{Training Set} & \textbf{Validation Set} & \textbf{Test Set} \\ \hline
Fold-1            & Halfmile, Lalor       & Brunswick               & Sudbury           \\
Fold-2            & Sudbury, Halfmile     & Lalor                   & Brunswick         \\
Fold-3            & Lalor, Brunswick      & Sudbury                 & Halfmile          \\
Fold-4            & Brunswick, Sudbury    & Halfmile                & Lalor             \\ 
Fold-5            & 80\% DN-1             & 20\% DN-1               & DN-2             \\ 
Fold-6            & DN-1                  & DN-2                    & LS             \\ 
\bottomrule\bottomrule
\end{tabular}
\end{table}

\subsection{Ablation Study}
We study the different components of MSSPN by removing STA/LTA map of RSN, Grad Loss of RSN, LMO processing, picking correction of labels, and the post-processing method, as shown in Tab. \ref{tab:ablation}. First, we study the necessity of input information. Specifically, removing the STA/LTA feature maps of RSN, represented by w/o STA/LTA in RSN, i.e., the input of RSN is a single cropped gather. Second, we study the loss function of RSN. Concretely, removing Grad Loss, denoted by w/o Grad Loss in RSN, i.e., we only use the BCE Loss to train RSN. Third, we study the necessity of using LMO correction to reduce input size, represented by w/o LMO, i.e., the input of RSN is cropped in time axis based on the minimum and maximum of the reference FAT obtained from RRVE. 
Finally, we test the improvement by the addition of the preprocessing and the post-processing. In the preprocessing, we focus on the picking correction process, denoted as w/o Pick Correction, which means that we directly input the manually picked input without any correction. Removing post-processing, denoted as w/o post-processing, i.e., the maximum indexes of each columns in the probability map of the RSN are marked as the final FAT. 

In the ablation experiments, the dataset splitting setting is Fold-4 in Tab. \ref{tab:fold-set}, and the hyperparameters of the model are the same as the parameter combinations in bold in Tab. \ref{tab:para-sel}. Each group of ablation experiments is repeated 10 times using different initialization network weights, and the best test result is recorded as the result of the current ablation model, as shown in Tab. \ref{tab:ablation}. 
The results show that the five ingredients in MSSPN are very reasonable settings and can improve effectively the quality of our picking results.

\begin{table}[h]
\centering
\caption{Test Results of The Ablation Study}
\label{tab:ablation}
\resizebox{0.95\linewidth}{!}{
\begin{tabular}{lccccc}
\toprule\toprule
\textbf{Model Variants} & \textbf{ACC@1 (\%)} & \textbf{ACC@3 (\%)} & \textbf{ACC@9 (\%)} & \textbf{MAE} & \textbf{RMSE} \\ \hline
w/o STA/LTA in RSN   & 87.90 & 89.02 & 98.12 & 0.72 & 11.97 \\
w/o Grad Loss in RSN   & 89.23 & 90.45 & 98.01 & 0.66 & 11.73 \\
w/o LMO              & 86.77 & 88.13 & 98.06 & 0.79 & 13.89 \\
w/o Pick-Correction  & 86.36 & 87.69 & 97.99 & 0.81 & 14.12 \\
w/o Post-Processing  & 91.41 & 92.57 & 98.43 & 0.52 & 10.42 \\ \hline
\textbf{Ours}        & \pmb{91.83} & \pmb{92.96} & \pmb{98.49} & \pmb{0.48}  & \pmb{8.58} \\ \bottomrule\bottomrule
\end{tabular}}
\end{table}

\begin{figure*}[ht!]
	\centering
	\subfloat[]{\includegraphics[height=4cm]{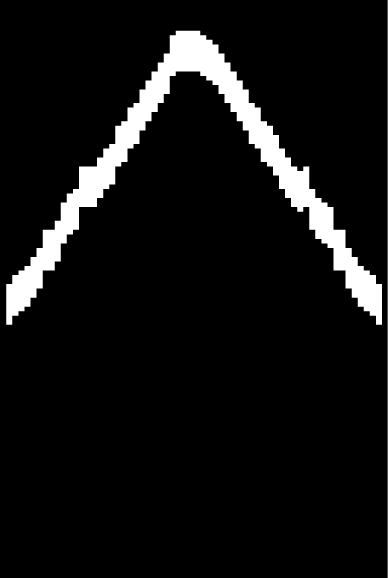}\label{fig: S1Mask}}
	\hfil
	\subfloat[]{\includegraphics[height=4cm]{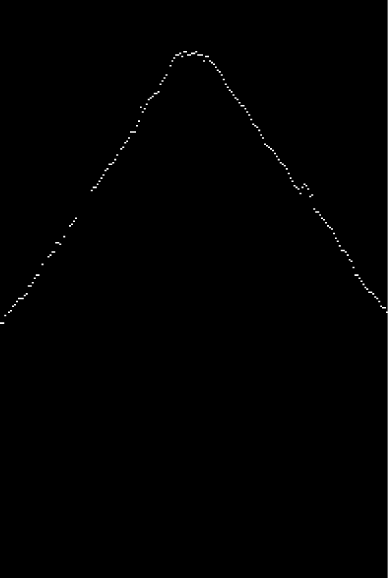}\label{fig: S3Mask}}
	\hfil
	\subfloat[]{\includegraphics[height=4cm]{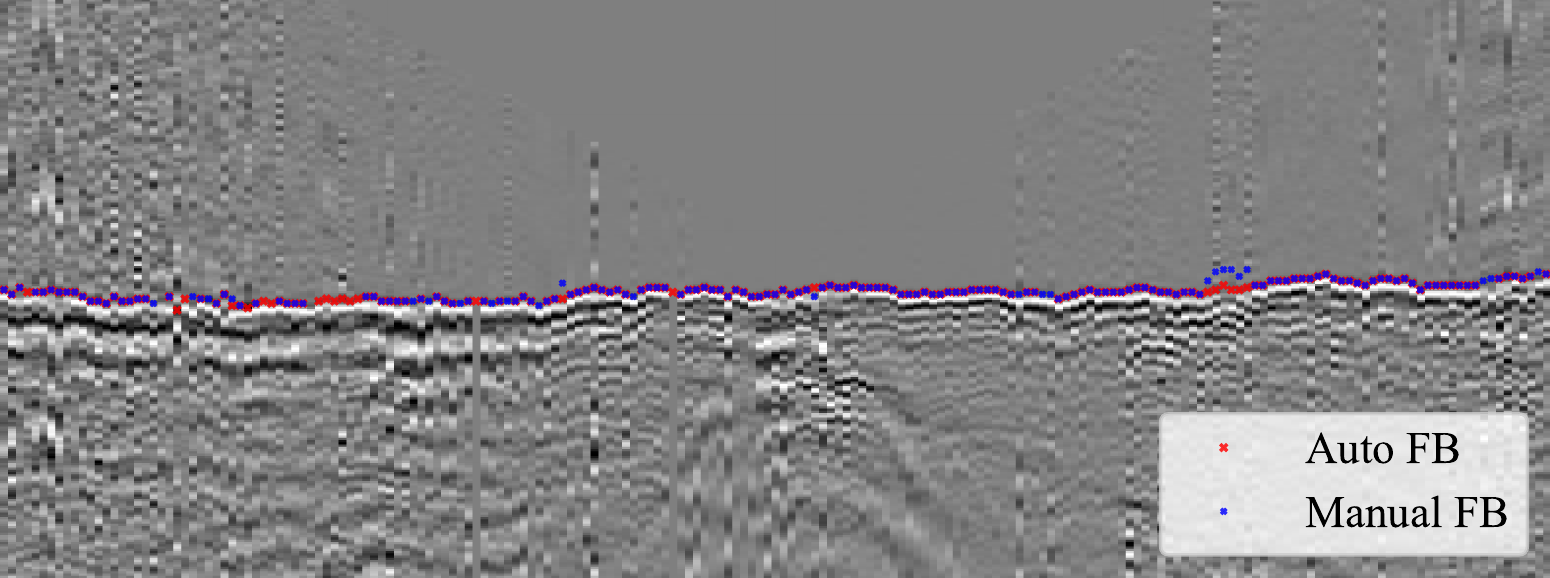}\label{fig: S3LMO}}
	\\
	\subfloat[]{\includegraphics[height=4cm]{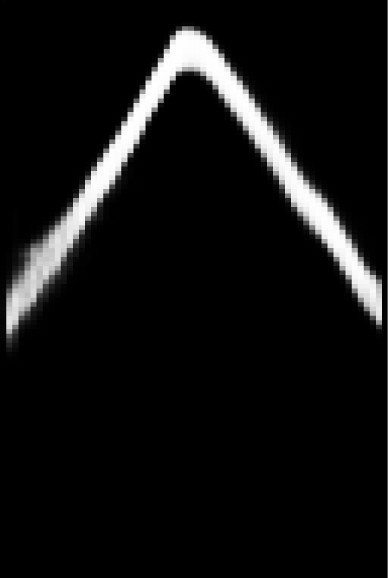}\label{fig: S1Pred}}
	\hfil
	\subfloat[]{\includegraphics[height=4cm]{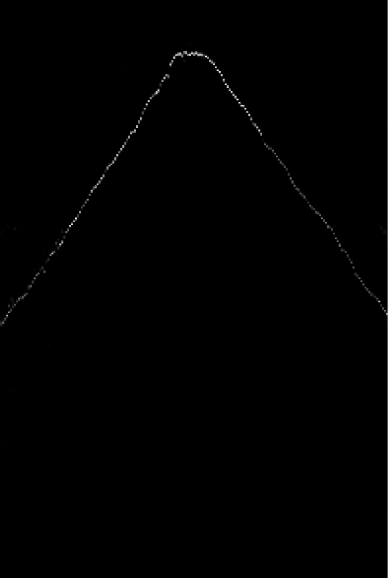}\label{fig: S3Pred}}
	\hfil
	\subfloat[]{\includegraphics[height=4cm]{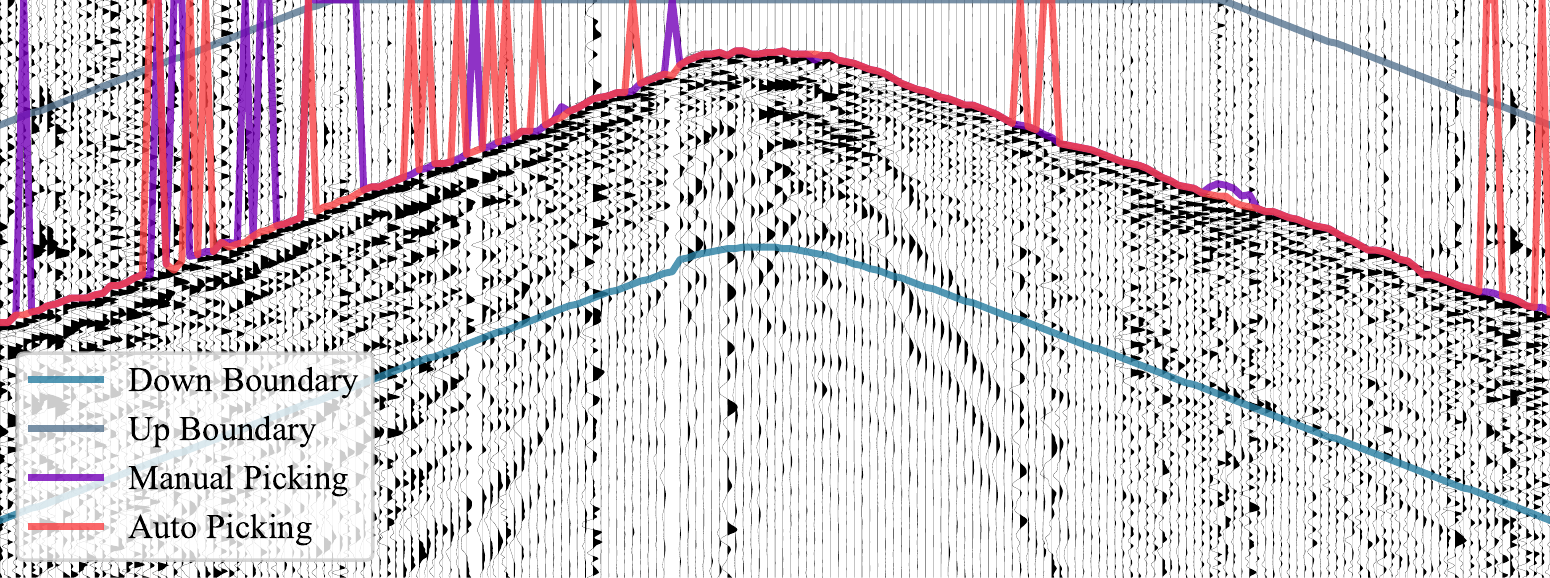}\label{fig: S3Final}}
	\\
	\subfloat[]{\includegraphics[height=3cm]{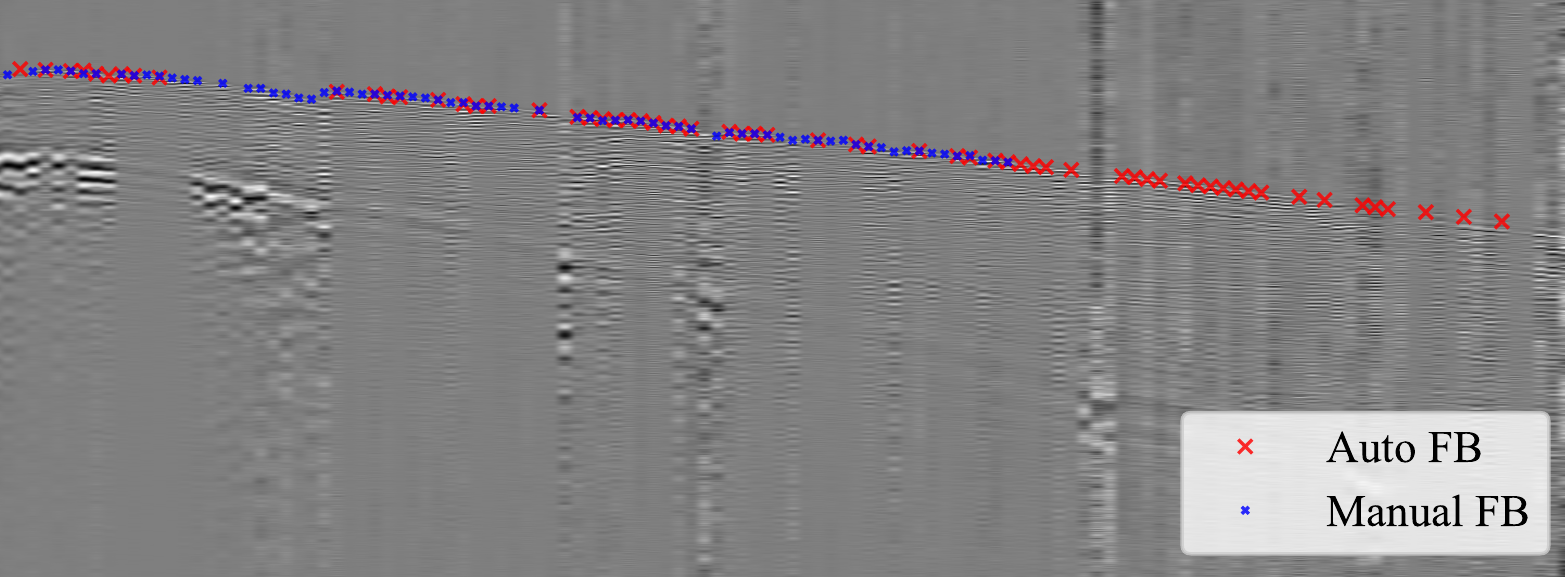}\label{fig: LaP}}
	\hfil
	\subfloat[]{\includegraphics[height=3cm]{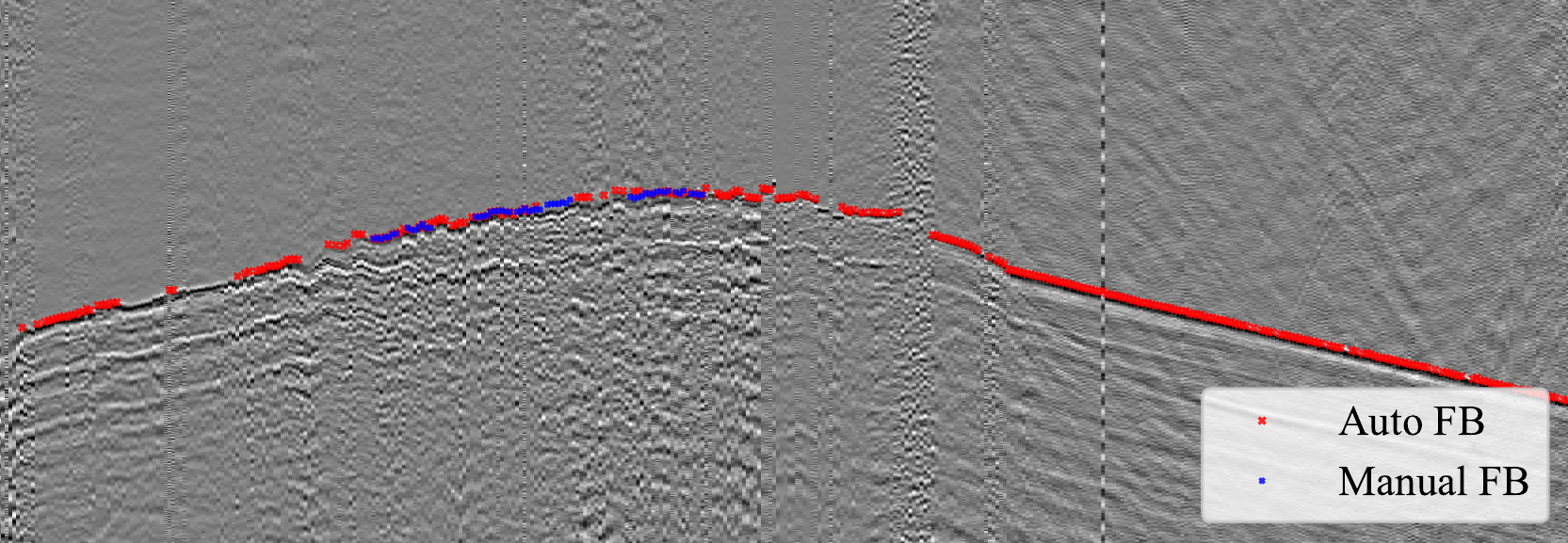}\label{fig: DNP}}
	\caption{(a)-(f) are a picking results of a medium SNR gather in Halfmile. (a), (d) and (b), (e) are ground-truths and prediction maps of CSN and RSN, respectively. (c) is a gather after LMO marked FAT also named First Break (FB) times. (f) is a wiggle plot of a gather with final FB and the boundaries of RSN input. (g) and (h) are medium SNR gathers with FB of Lalor and DN-2, respectively.}
	\label{fig: PickProcess}
\end{figure*}
\subsection{Results}
After evaluating the necessity of main ingredients in our proposed MSSPN, in this section, we will introduce more experiment results about the comparative experiment between MSSPN and current popular models, and the test results on more datasets.
\subsubsection{Comparative Experiment}
To verify the advantages of our model in generalization across worksites, we compare two current popular FAT automatic picking methods: U-Net-based benchmark\cite{P2021Neurips} and nested U-Net-based method\cite{zhang2020first}. The training settings of these two methods all refer to the settings in the original paper, and we choose Fold-4 (medium SNR) and Fold-6 (low SNR) in Tab. \ref{tab:fold-set} to evaluate these models. As shown in Tab. \ref{tab:comp-ep}, on the Lalor dataset with medium SNR, we have achieved 91.83\% of the completely picking accuracy and 0.48 of MAE. Even more surprising is that compared to the baseline model, our picking results on the low SNR dataset (LS) are significantly improved, which is unattainable by these two one-stage segmentation models.

\begin{table}[h]
\centering
\caption{Test Results of The Comparative Experiment}
\label{tab:comp-ep}
\resizebox{0.95\linewidth}{!}{
\begin{tabular}{lcccccc}
\toprule\toprule
\multicolumn{1}{c}{\multirow{2}{*}{\textbf{Model}}} & \multicolumn{3}{c}{Lalor}      & \multicolumn{3}{c}{LS}        \\ \cline{2-7} 
\multicolumn{1}{c}{}    & ACC@1 (\%) & ACC@3 (\%) & MAE   & ACC@1 (\%) & ACC@3 (\%) & MAE  \\ \hline
U-Net(Benchmark)        & 76.30     & 80.00     & 124.00 & 8.28      & 34.93     & 16.15 \\
Nested U-Net            & 77.05     & 87.31     & 2.53  & 17.16     & 38.90     & 46.54 \\ \hline
\pmb{Ours}              & \pmb{91.83}     & \pmb{92.96}     & \pmb{0.48}   & \pmb{69.43}     & \pmb{70.07}     & \pmb{6.05}  \\ \bottomrule\bottomrule
\end{tabular}}
\end{table}

\subsubsection{Performance of MSSPN}
In this subsection, we will introduce the visual results and quantitative results of MSSPN. 
First, a visual picking processing of a gather with the medium SNR in Halfmile is shown in Fig. \ref{fig: S1Mask} - \ref{fig: S3Final}, and the setting of dataset is Fold-2 in Tab. \ref{tab:fold-set}. 
From Fig. \ref{fig: S1Mask} and \ref{fig: S1Pred}, we can see that ground-truth (Fig. \ref{fig: S1Mask}) guides CSN to segment the fuzzy region of FAT. 
Next, the local velocity $v$ and zero-offset traveltime $t_0$ in LMO can be estimated by RANSACwVC based on segmentation map of CSN, and the estimation results are $v = 5097 m/s$ and $t_0 = 0.0251$. 
Then, we calculate the reference FAT using Eq. \ref{PredT}. We let the length of the confidence interval be 128, and then we obtain the gather after LMO with height = 256 shown in Fig. \ref{fig: S3LMO}. 
Next, the gather after LMO is cropped by column and is input into RSN, and the spliced segmentation map of RSN is shown in Fig. \ref{fig: S3Pred}. From Fig. \ref{fig: S3Mask}, we can see that the ground-truth of RSN is fine segmentation, and the smaller input of RSN greatly alleviates unstable segmentation results caused by sparse labels as shown in Fig. \ref{fig: S3Pred}. 
The final FAT is shown in Fig. \ref{fig: S3LMO} and \ref{fig: S3Final}, and the interior of the boundaries in Fig. \ref{fig: S3Final} corresponds to the gather after LMO (Fig. \ref{fig: S3LMO}). From the final FAT, we can find that some automatic picking results surpass manual picking results, and the errors of manual picking may be caused by strong noise near FAT. 
Second, we also show the picking results of two gathers with the medium SNR in Lalor and DN-2 in Fig. \ref{fig: LaP} and \ref{fig: DNP}. From Fig. \ref{fig: S3Final}, \ref{fig: LaP}, and \ref{fig: DNP}, we can find that the traces with far offset distances are often accompanied by stronger noises. Therefore, these traces are not picked manually in general. Surprisingly, MSSPN has strong anti-manufacturing ability and can also achieve good performance on the traces with far offset distances. 

In Tab. \ref{tab:fold-set}, we set four-folds cross-validated datasets for open datasets as the same as the setting in \cite{P2021Neurips} and two-folds datasets for BGP datasets to test MSSPN quantitatively. It is worth to note that our test dataset and training dataset are different worksites, so this test fully simulates the actual prediction scenario. The test results are shown in Tab. \ref{tab:MSSPN}, where we also calculate the improvement of the ACC@1 metric compared to the benchmark\cite{P2021Neurips}, and show it in the last column. As shown in Tab. \ref{tab:MSSPN}, We can achieve full correct rates (ACC@1) of over 98\% on high SNR datasets (e.g. Brunswick) and over 90\% on medium SNR datasets (e.g. Lalor). Even more gratifying is that MSSPN achieves a full correct rates of nearly 70\% across worksites on low SNR dataset (LS). Moreover, according to the last column of Tab. \ref{tab:MSSPN}, compared with the baseline model\cite{P2021Neurips}, MSSPN has a significant improvement, especially on the low SNR dataset (LS). 

\begin{table}[h]
\centering
\caption{Test results of MSSPN}
\label{tab:MSSPN}
\resizebox{\linewidth}{!}{
\begin{tabular}{lcccccc}
\toprule\toprule
\textbf{Test Set} & \textbf{ACC@1 (\%)} & \textbf{ACC@3 (\%)} & \textbf{ACC@9 (\%)} & \textbf{MAE} & \textbf{RMSE} & \textbf{Improvement (\%)} \\
\hline
Sudbury   & 95.63 & 95.67 & 97.80 & 0.88 & 6.62 & +30.82\\
Brunswick & 98.05 & 98.27 & 99.59 & 0.13 & 6.48 & +11.93\\
Halfmile  & 94.88 & 95.06 & 99.34 & 0.33 & 14.75 & +13.22\\
Lalor     & 91.83 & 92.96 & 98.60 & 0.48 & 8.58 & +20.35\\
DN-2      & 97.30 & 98.02 & 99.45 & 0.16 & 8.42 & +10.58\\
LS        & 69.42 & 70.07 & 71.83 & 6.05 & 65.21 & +738.52\\
\bottomrule\bottomrule
\end{tabular}}
\end{table}

\subsubsection{Fine-tuning Test on Dataset with Low SNR}
\begin{figure}[ht!]
	\centering
	\subfloat[]{\includegraphics[width=0.9\linewidth]{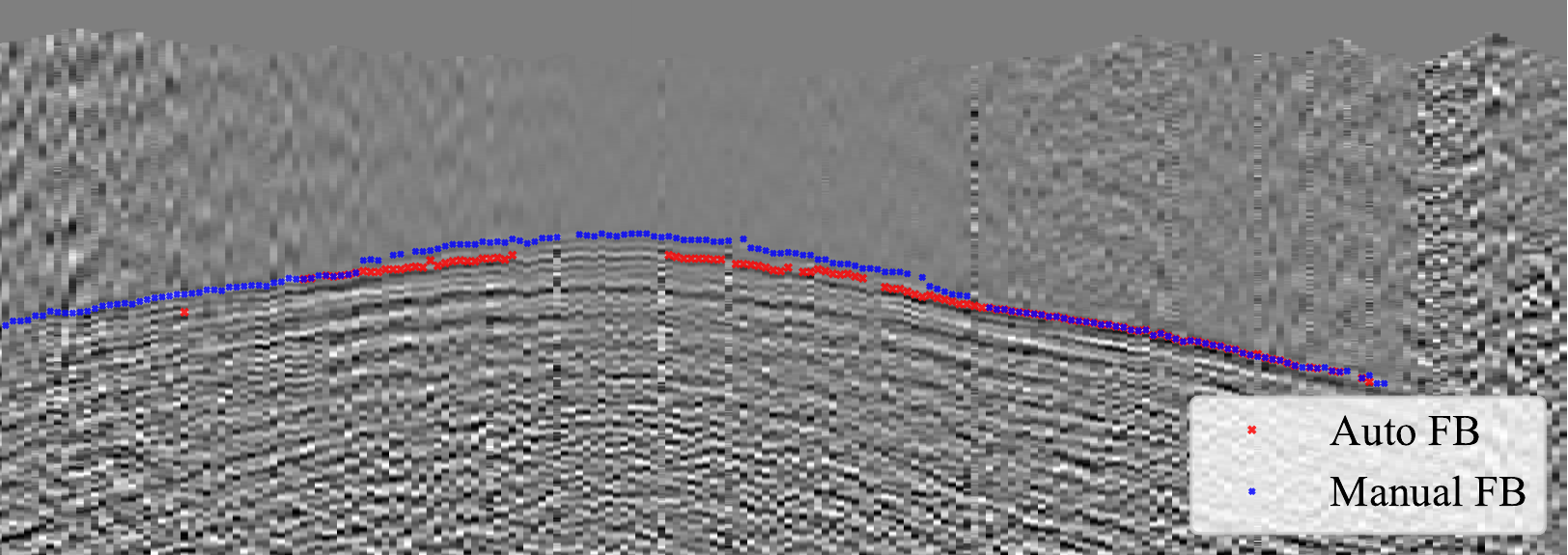}\label{fig: LS-UT}}\\
	\subfloat[]{\includegraphics[width=0.9\linewidth]{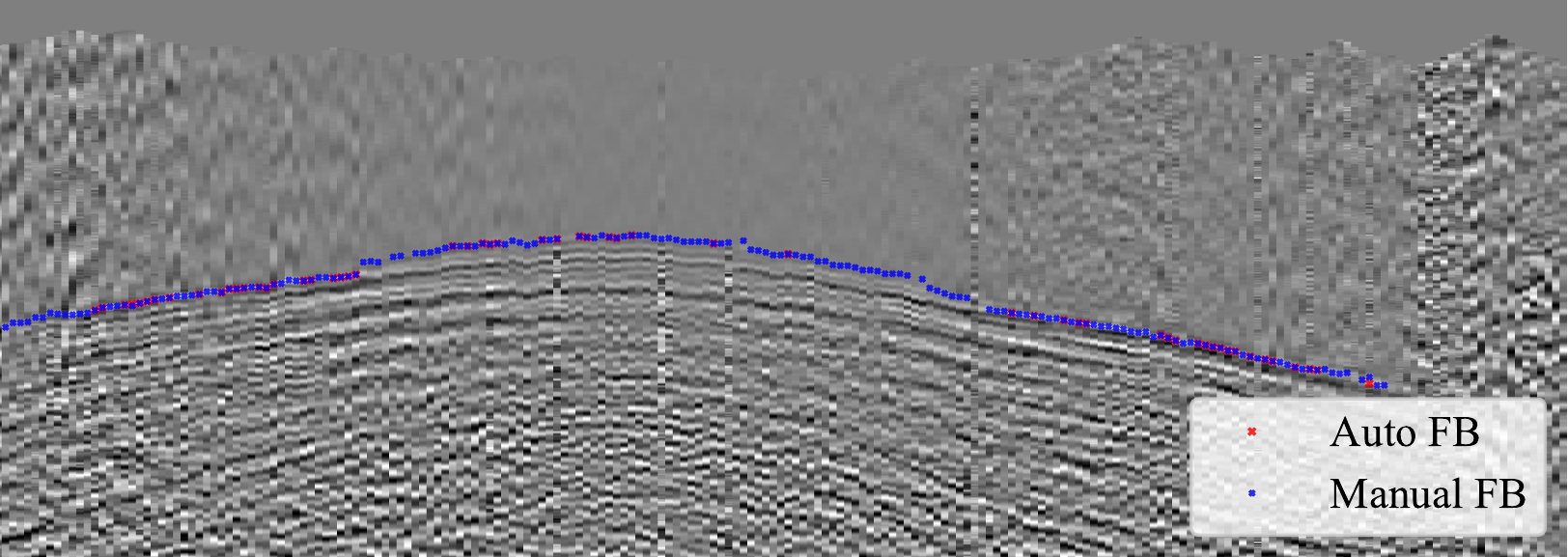}\label{fig: LS-T}}
	\caption{A low SNR gather with FB of LS. (a) The picking result before fine-tuning. (b) The picking result after fine-tuning.}
	\label{fig: FineTuning}
\end{figure}

To further evaluate the potential of MSSPN in case of low SNR, we try a fine-tune training method, which is a method of few-shot learning. Concretely, we let the trained model based on Fold-6 be a pretrained model, and training set of fine-tuning learning is a part of dataset LS with a volume of 512 shots. Except for the training batch size adjusted to 16, other training settings are consistent with Tab. \ref{tab:para-sel}. Next, we test the fine-tuned model on the remaining samples of LS. The model before fine-tuning incorrectly identifies FAT in the middle part of the gather due to weak signals not being identified as shown in Fig. \ref{fig: LS-UT}, and the model after fine-tuning perceives these shallow weak signals, and removes the pick results with low confidence as shown in Fig. \ref{fig: LS-T}. Quantitatively, the ACC@1 of the fine-tuned model is 88.06\%, an increase of 26.83\% compared to the model without fine-tuning (69.43\%). Therefore, we verify that the fine-tuned-trained MSSPN model can achieve better picking accuracy on low SNR dataset.

\section{Conclusion}
In this paper, we propose a multi-stage segmentation picking network named MSSPN, in which we decompose the first arrival time picking task into four simple tasks to improve the picking accuracy and the transferability of the model. The following three conclusions can be drawn from various experiments. (1) Our picking framework can solve the hard problem of picking across worksites and produce more robust picking results with higher accuracy comparing to current popular automatic picking method, especially in case of low SNR. (2) The use of LMO can better normalize the input of semantic segmentation-based methods, so that the transferability of networks can be improved greatly. (3) Fine-tuning learning can learn more information of a new data domain, so that the picking accuracy is further improved on the dataset with low SNR.  

\section*{Acknowledgment}
The author would like to thank Mr Pierre-Luc St-Charles from Applied Machine Learning Research Team
Mila, Québec AI Institute and Mr Zhenbo Guo from Bureau of Geophysical Prospecting (BGP) for providing the open datasets and BGP datasets, respectively. 
\ifCLASSOPTIONcaptionsoff
  \newpage
\fi

\bibliographystyle{IEEEtran}
\bibliography{reference}

\begin{IEEEbiography}[{\includegraphics[width=0.9in,height=1.25in,clip]{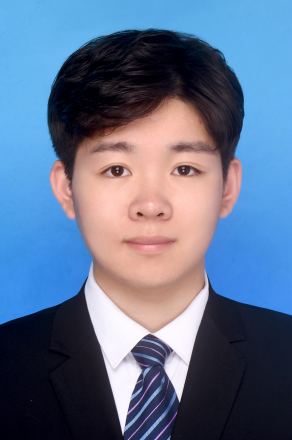}}]{Hongtao~Wang} is currently pursuing the Ph.D. degree in statistics with the School of Mathematics and Statistics, Xi’an Jiaotong University, Xi’an, China. His research interests include Bayesian statistics and deep learning.
\end{IEEEbiography}
\begin{IEEEbiography}[{\includegraphics[width=0.9in,height=1.25in,clip]{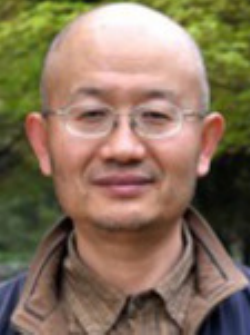}}]{Jiangshe~Zhang} was born in 1962. He received the M.S. and Ph.D. degrees in applied mathematics from Xi'an Jiaotong University, Xi'an, China, in 1987 and 1993, respectively, where he is currently a Professor with the Department of Statistics. He has authored and co-authored one monograph and over 80 conference and journal publications on robust clustering, optimization, short-term load forecasting for the electric power system, and remote sensing image processing. His current research interests include Bayesian statistics, global optimization, ensemble learning, and deep learning.
\end{IEEEbiography}

\begin{IEEEbiography}[{\includegraphics[width=0.9in,height=1.25in,clip]{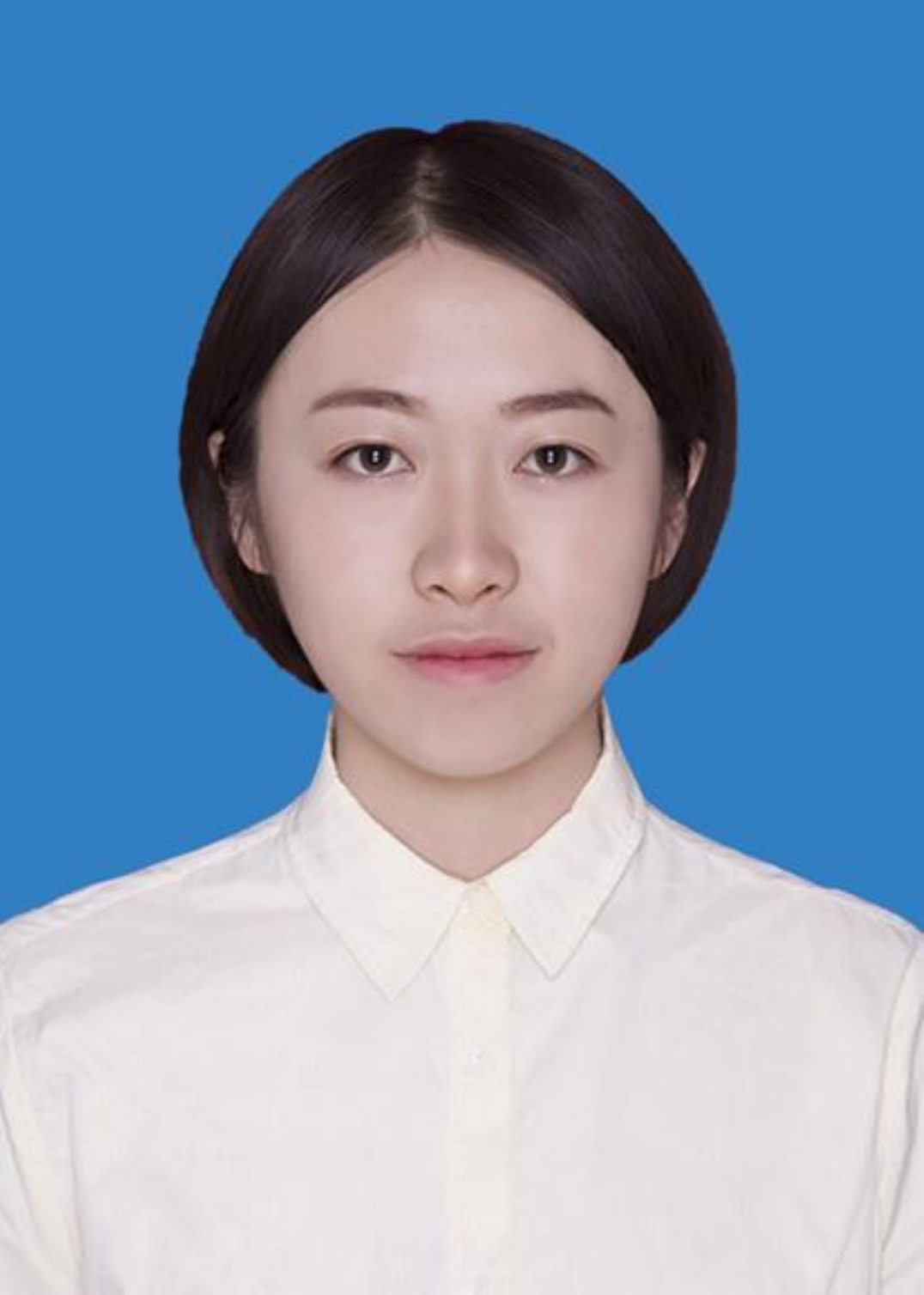}}]{Xiaoli~Wei} is currently pursuing the Ph.D. degree in statistics with the School of Mathematics and Statistics, Xi’an Jiaotong University, Xi’an, China.
Her research interests include seismic data reconstruction, deep learning and uncertainty estimation.
\end{IEEEbiography}

\begin{IEEEbiography}[{\includegraphics[width=0.9in,height=1.25in,clip]{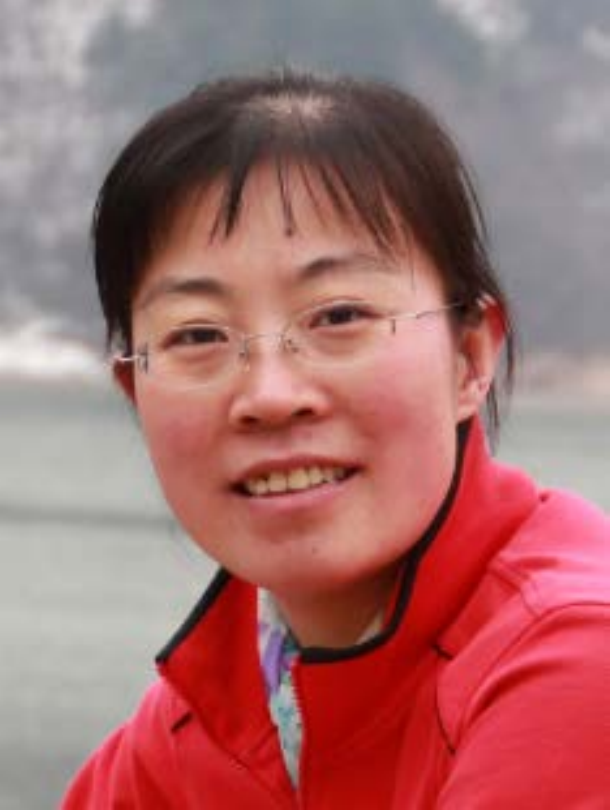}}]{Chunxia~Zhang} received her Ph.D degree in Applied Mathematics from Xi'an Jiaotong University, Xi'an,  China, in 2010. Currently, she is a Professor in School of Mathematics and Statistics at Xi'an Jiaotong University. She has authored and coauthored about 30 journal papers on ensemble learning techniques, nonparametric regression, etc. Her main interests are in the area of ensemble learning, variable selection and deep learning.
\end{IEEEbiography}

\begin{IEEEbiography}
[{\includegraphics[width=0.9in,height=1.25in,clip]{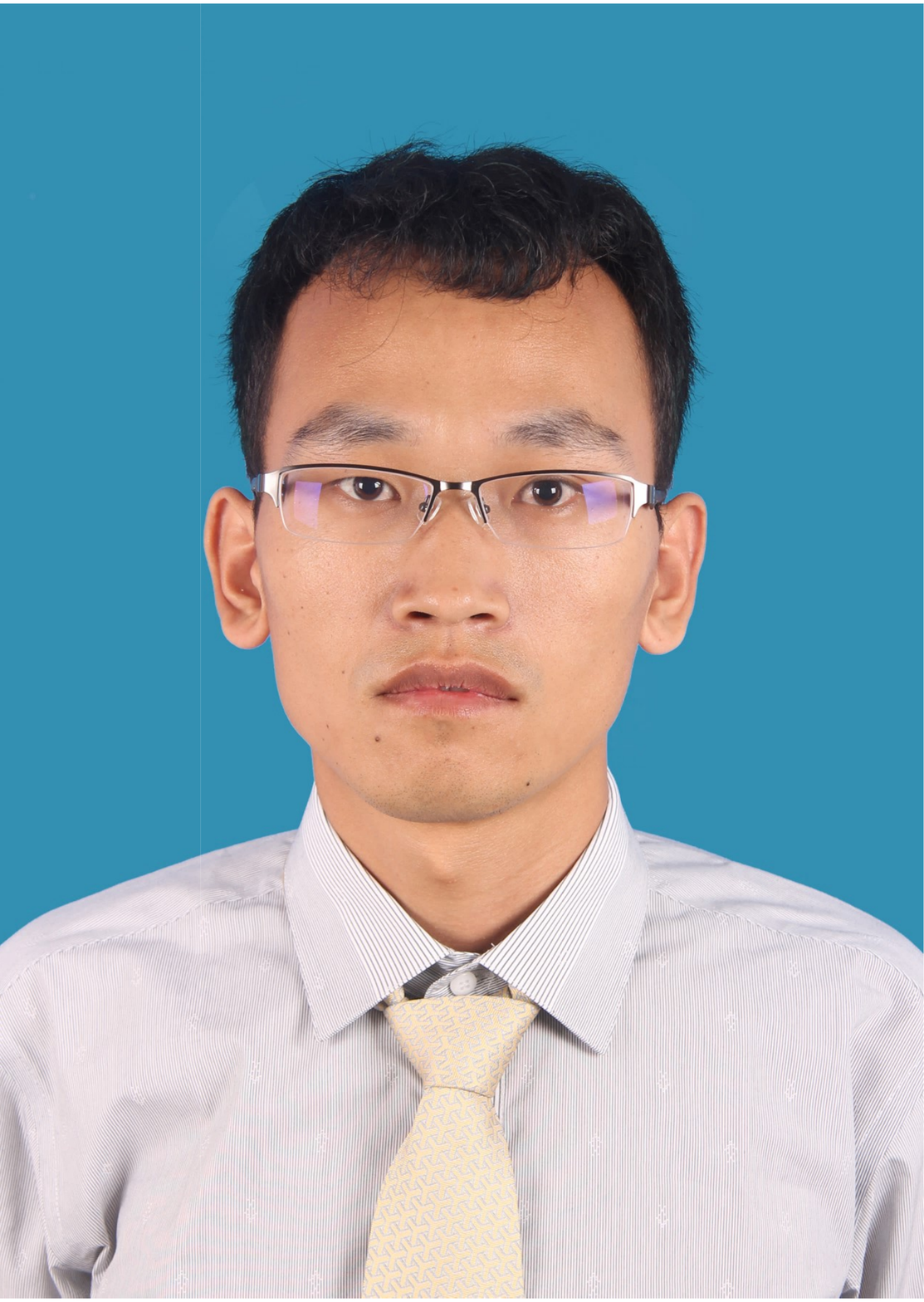}}]{Zhenbo~Guo} is currently a senior geophysical engineer at R\&D Center, BGP, CNPC, China. His research interests include seismic inversion and application of deep learning in seismic data processing.
\end{IEEEbiography}

\begin{IEEEbiography}[{\includegraphics[width=0.9in,height=1.25in,clip]{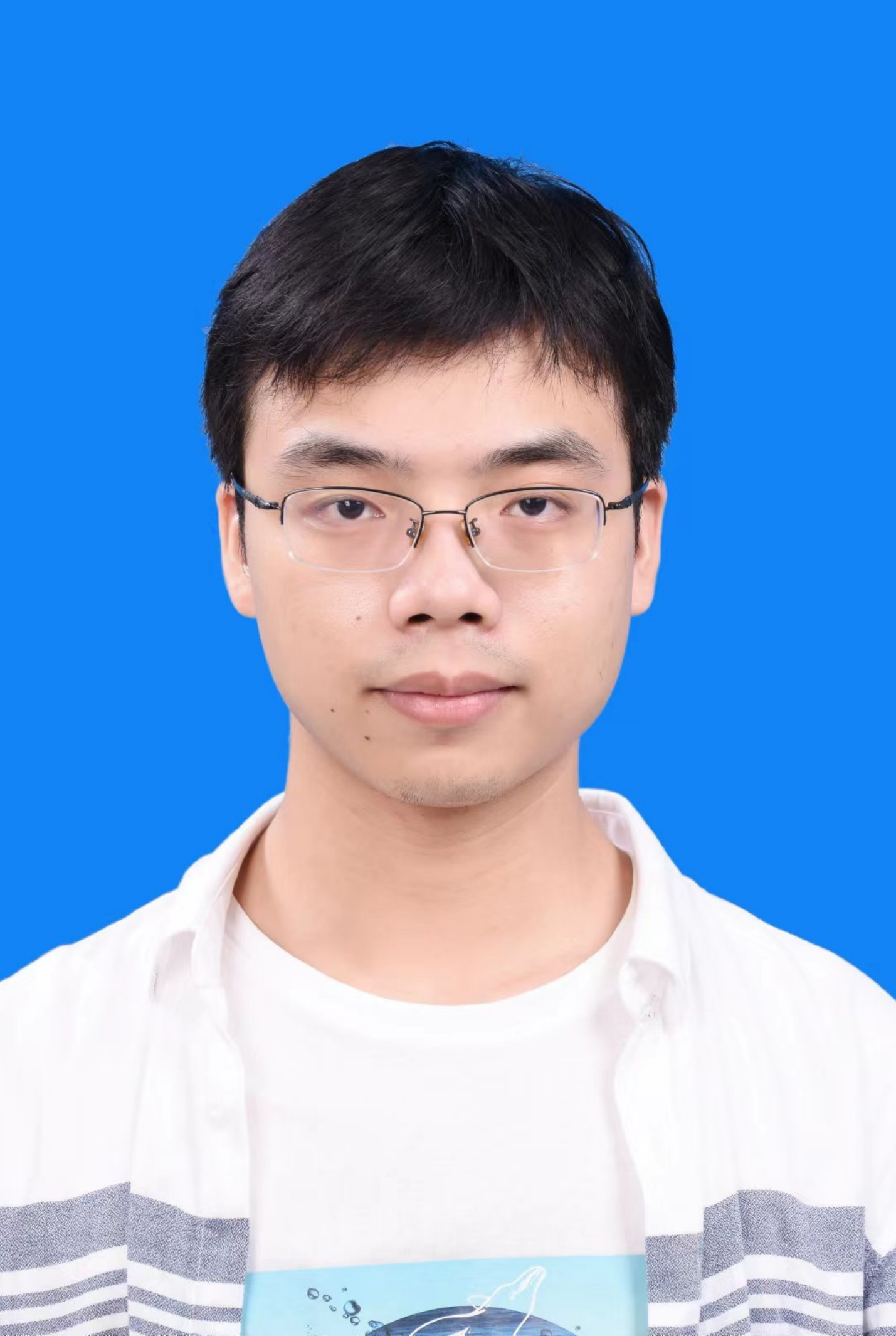}}]{Li~Long} is a Ph.D. candidate in statistics with the School of Mathematics and Statistics, Xi’an Jiaotong University, Xi’an, China. His research interests include reinforcement learning.
\end{IEEEbiography}

\begin{IEEEbiography}
[{\includegraphics[width=0.9in,height=1.25in,clip]{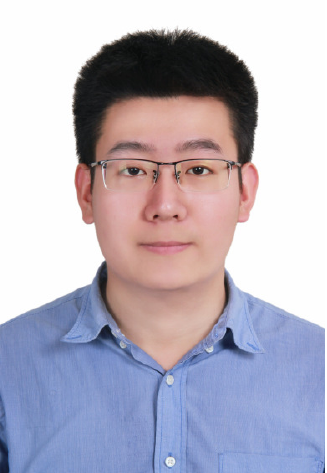}}]{Yicheng~Wang} is now pursuing for the Ph.D. degree in statistics at Xi'an Jiaotong University. His research interests include low-level computer vision, generative model and out-of-distribution detection.
\end{IEEEbiography}

\end{document}